\documentclass{article}
\usepackage{todonotes}

    \usepackage[preprint]{neurips_2025}

\usepackage[utf8]{inputenc} 
\usepackage[T1]{fontenc}    
\usepackage{hyperref}       
\usepackage{url}            
\usepackage{booktabs}       
\usepackage{amsfonts}       
\usepackage{nicefrac}       
\usepackage{microtype}      
\usepackage{xcolor}         
\usepackage{graphicx}
\usepackage{amsmath}
\usepackage{subcaption}
\usepackage{authblk}

\title{Predicting Reaction Time to Comprehend Scenes with Foveated Scene Understanding Maps }

\author{
  \textbf{Ziqi Wen} \textsuperscript{1}  \quad\ \textbf{Jonathan Skaza}  \textsuperscript{2}  \quad\  \textbf{Shravan Murlidaran} \textsuperscript{3} \\
 \quad\ \textbf{William Y. Wang} \textsuperscript{1} \quad\ \textbf{Miguel P. Eckstein}\textsuperscript{1,2,3}

\textsuperscript{1} Department of Computer Science, University of California, Santa Barbara\\
\textsuperscript{2} Graduate Program in Dynamical Neuroscience, University of California, Santa Barbara \\
\textsuperscript{3} Department of Psychological and Brain Sciences, University of California, Santa Barbara \\
  \tt\small{\{ziqiwen, skaza, smurlidaran, migueleckstein\}@ucsb.edu \quad william@cs.ucsb.edu  } \\
}

\begin{document}

\maketitle

\begin{abstract}

Although models exist that predict human response times (RTs) in tasks such as target search and visual discrimination, the development of image-computable predictors for scene understanding time remains an open challenge. Recent advances in vision-language models (VLMs)—which can generate scene descriptions for arbitrary images—combined with the availability of quantitative metrics for comparing linguistic descriptions, offer a new opportunity to model human scene understanding. We hypothesize that the primary bottleneck in human scene understanding and the driving source of variability in response times across scenes is the interaction between the foveated nature of the human visual system and the spatial distribution of task-relevant visual information within an image. Based on this assumption, we propose a novel image-computable model that integrates foveated vision with VLMs to produce a spatially resolved map of scene understanding as a function of fixation location (Foveated Scene Understanding Map, or F-SUM), along with an aggregate F-SUM score. This metric correlates with average ($N=17$) human RTs ($r=0.47$) and number of saccades ($r=0.51$) required to comprehend a scene (across 277 scenes). The F-SUM score also correlates with average ($N=16$) human description accuracy ($r=-0.56$) in time-limited presentations. These correlations significantly exceed those of standard image-based metrics such as clutter, visual complexity, and scene ambiguity based on language entropy. Together, our work introduces a new image-computable metric for predicting human response times in scene understanding and demonstrates the importance of foveated visual processing in shaping comprehension difficulty.

\end{abstract}

\section{Introduction \& Former Research}


There is a long scientific history~\citep{donders1969speed,pieron1913ii} of trying to understand and model how visual information influences response (reaction) times (RT) in perceptual tasks. The investigations have progressed from the detection of simple visual and auditory stimuli \citep{donders1969speed}, identification of letters and words \citep{cattell1885inertia}, search for targets in multi-element arrays 
\citep{neisser1967cognitive}, object categorization \citep{vanrullen2001time},  discrimination of faces \citep{fraser1990reaction}, and scene gist recognition (\citep{joubert2007processing}) .    

A subset of these studies has focused on understanding how response times vary with basic stimulus properties such as the intensity of a flash,  the target/distractor similarity in visual search \citep{duncan1989visual}, the direction of motion \citep{dzhafarov1993detection} or more high-level semantic properties such as contextual congruency of objects \citep{mack2008object}.

Investigators have also developed computational models that operate on input images (image-computable models) to generate a score that correlates with human response times \citep{mirzaei2013predicting, balakrishnan2014comparative, woods2015factors}. For example, image computable models that estimate edge density, entropy have been used to predict recognition time \citep{mack2004computational,rosenholtz2005feature, rosenholtz2007measuring}. Models that quantify the crowding of visual information or features have been proposed as predictors of clutter judgments or target search times \citep{bravo2008scale, deza2016can}.



Human perceptual experience in everyday life involves more than basic visual tasks (detection, identification, categorization of objects and scenes); it consists of comprehending the complex human and social behaviors (e.g., a child throwing a ball to have the dog run to catch it).   Until recently, numerous obstacles made it unfeasible to model human reaction times to understand a scene.  First, there were no adequate quantitative methods and metrics to assess whether humans understand a scene.  The common categorical decisions in perceptual psychology used for letter, scene, and object recognition cannot capture the richness of the perceptual experience of comprehending a scene.  Second, image-computable models could not generate a scene description from arbitrary scenes.   

The rapid improvement of vision-language models (VLMs) over the last five years has resulted in models that can generate scene descriptions for any image input \citep{radford2021learning, li2022blip,liu2023visual,achiam2023gpt,lu2024deepseek}. Similarity metrics based on language embeddings allow quantitative evaluations of the similarity of a human description of a scene to a gold standard description \citep{cer2018universal, zhang2019bertscore, reimers2019sentence, li2023towards, lee2024nv}. These new powerful VLM tools can be used to predict the time that a human requires to comprehend a scene. 

\begin{figure}[ht]
    \centering
    \includegraphics[width=0.8\linewidth]{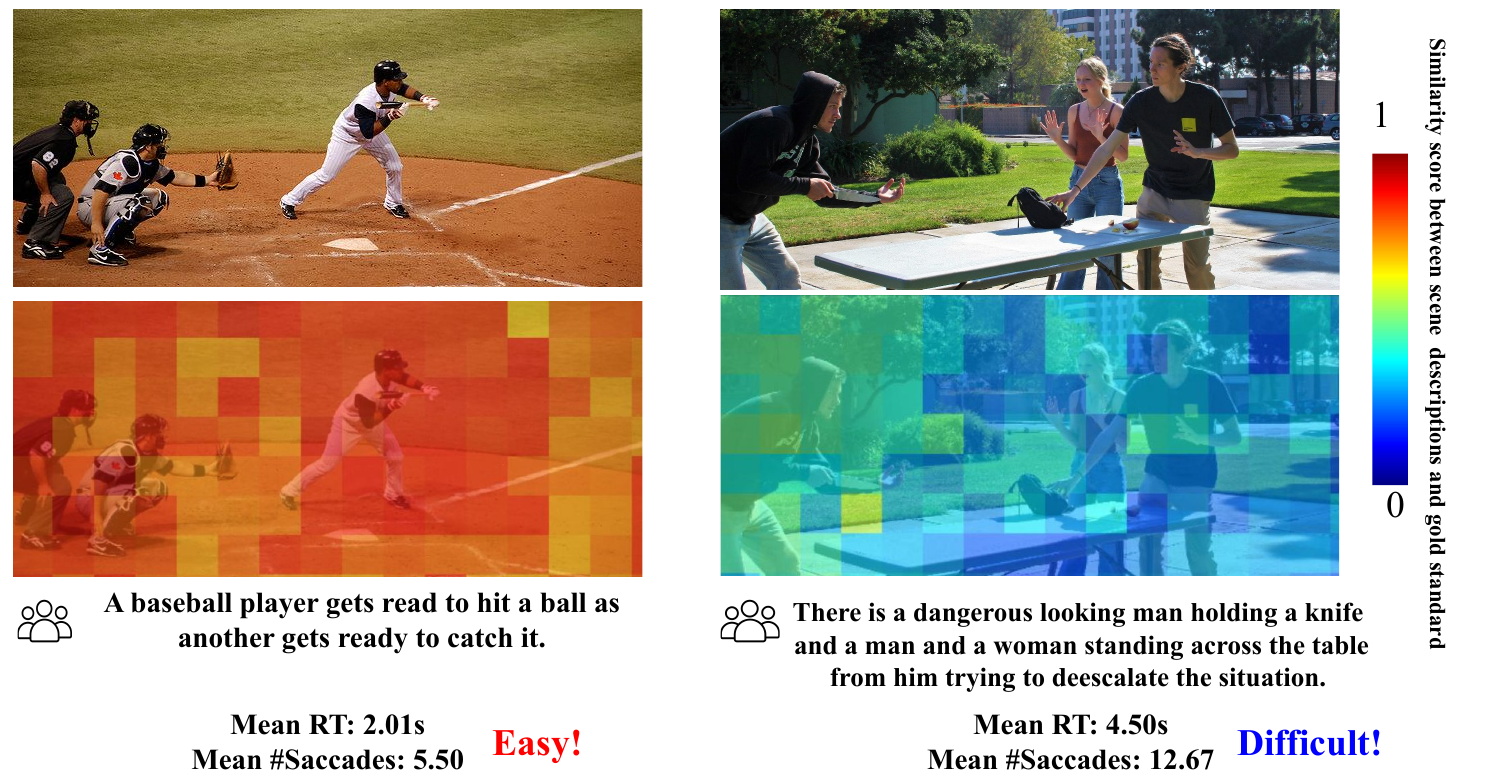}
    \caption{Comparison between low-effort and high-effort scene understanding. For the left scene, all the critical elements to understand the scene (baseball players) are easily identifiable in the visual periphery  As a result, participants can quickly comprehend the scene regardless of fixation location, which is reflected in F-SUM; looking anywhere can obtain descriptions similar to the gold standards. In contrast, the right scene represents a high-effort scene, key elements (e.g. knife, facial expression, bag) are spatially distributed, not easily identified in the visual periphery requiring multiple fixations. Consequently, the VLM single fixation descriptions diverge from the gold standards.}
    \label{fig:intro_examples}
\end{figure}

Here, we hypothesize that the main bottleneck in human scene understanding and the driving variability in response time across scenes is the interaction between the foveated nature of the human visual system (the fact that humans see high spatial detail where they are looking and have degraded vision away from the fixation point, the visual periphery) and the spatial distribution across the image of the visual information critical to comprehending the scene.  For example, if understanding a scene involves processing details not identifiable in the visual periphery, then comprehending the scene will require eye movement exploration and increase the human response times (e.g., right image, Fig.\ref{fig:intro_examples}). In contrast,  if understanding the scene involves large objects that are easily identified in the visual periphery or requires fixating a single object in the scene, then human response times to comprehend the scene will be shorter (e.g., left scene in Fig.\ref{fig:intro_examples}).  

Based on this premise, we combine VLMs with simulated human foveation to generate a predictive map of how fixating at different points on an image influences the VLM's description of the scene (Foveated Scene Understanding Map, F-SUM, see maps in Fig.\ref{fig:intro_examples}). An aggregate across the F-SUM is used as a score that predicts human response times to comprehend the scene.  We assess the ability of the F-SUM score to predict human response times and the number of eye movements required to comprehend the scene.  We also evaluate the F-SUM's score's ability to predict how well humans understand scenes when they have a limited number of saccades (rapid eye movements) to comprehend the scene.

We compare the newly proposed F-SUM score to previously proposed image-computable measures of image clutter \citep{mack2004computational, rosenholtz2005feature} and image complexity \citep{feng2022ic9600, mahon2024minimum}.  These measures capture important lower-level and higher-level image properties but do not consider which parts of the image contribute to the understanding of the scene.  In addition, we compare a metric that measures the uncertainty in the scene descriptions \citep{malinin2020uncertainty, kuhn2023semantic,nikitin2024kernel} without incorporating the interaction of the image information with the foveated human visual system.  As a final control, we compare the F-SUM score to a multimodal large language model (GPT-4o) prediction of the estimated relative response times of the different scenes.






\section{Foveated Scene Understanding Map (F-SUM)}

\begin{figure}[ht]
    \centering
    \includegraphics[width=\linewidth]{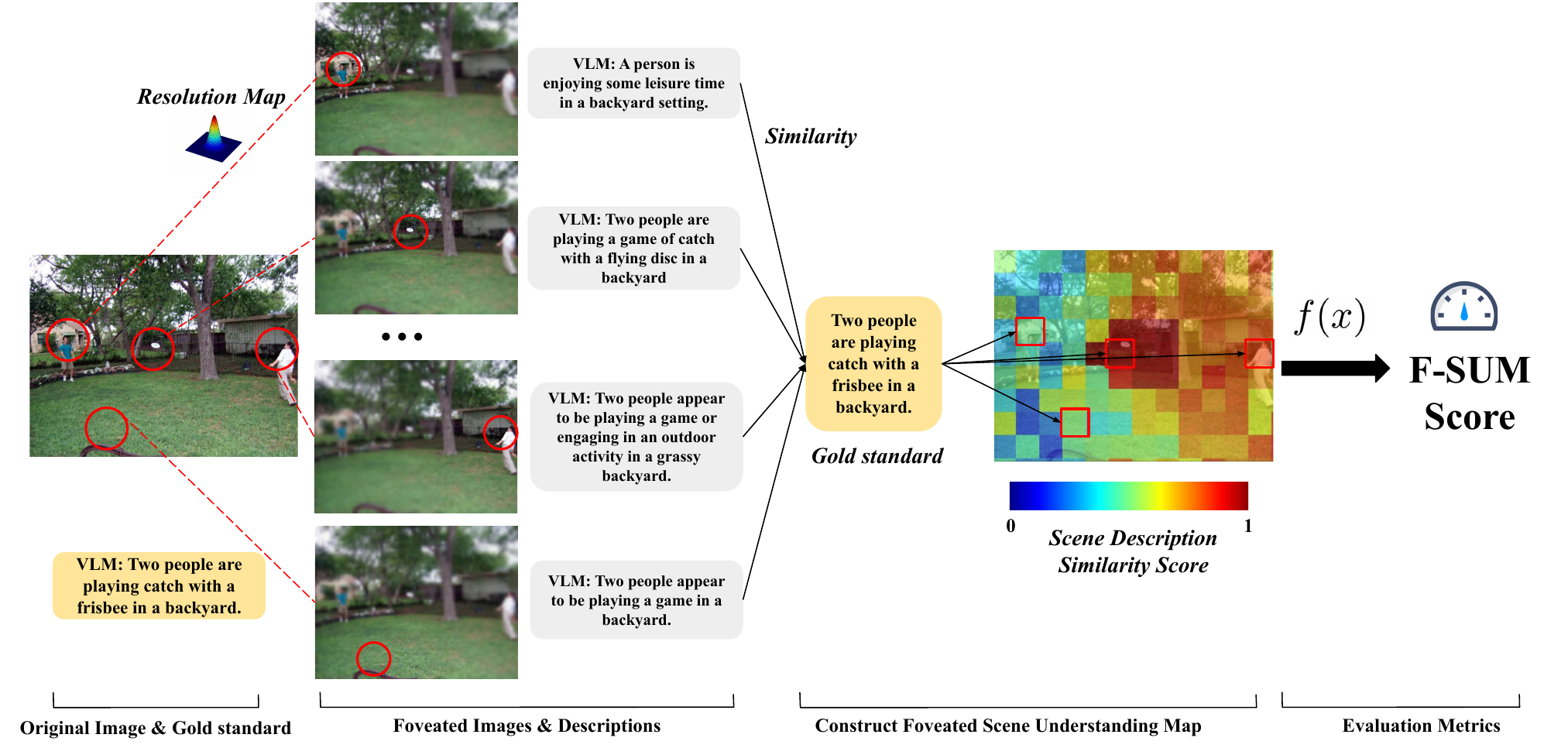}
    \caption{Overview of the F-SUM method. (1) Apply a VLM to obtain the description for the original images, deemed the gold standard description of the scene. (2) Use a foveation model to create an array of foveated images across many possible points of fixation, and apply the same VLM to return the description for each foveated rendering. (3) Construct the F-SUM based on the similarity between the gold standard and descriptions for the foveated renderings. (4) Apply the metrics using the constructed F-SUM to get a final score that measures the difficulty of understanding the scene. }
    \label{fig:method}
\end{figure}

Here, we present an overview of our computational approach to predict scene comprehension time in humans. To begin, we process the original scene through a VLM to obtain a description, serving as a ``gold standard'' representing complete scene understanding without foveation constraints. Subsequently, we apply a foveation model to generate image variants mimicking different fixation points and re-use the same VLM to extract descriptions for these foveated scenes (Sec.~\ref{sec:Fov}). We then construct the F-SUM by measuring similarity between the gold standard description embedding and descriptions from the foveated scenes (Sec.~\ref{sec:F-SUM}). Finally, we apply our metrics to the F-SUM matrix to aggregate across the map and quantify scene comprehension difficulty score  (Sec.~\ref{sec:metrics}). Fig.~\ref{fig:method} illustrates the complete pipeline. Each step is explained in more detail below.

 
\subsection{Scene Descriptions for Foveated Images}\label{sec:Fov}

To construct the F-SUM, we sample 108–136 fixation points\footnote{The exact number depends on image dimensions/aspect ratio.} across each scene using a uniform grid. At each point, we apply a foveation model that simulates the fall-off in human visual acuity with eccentricity \citep{perry2002gaze, jiang2015salicon}.  A VLM then generates a description for each foveated image, capturing the information accessible under that fixation. In our experiments, we show two variations of implementation, using a more powerful closed-source model (GPT-4o) or a smaller but efficient open-source model (Ovis2-8B)\citep{lu2024ovis}.
Each description serves as a proxy for what a human observer might perceive at that location.

The foveation process applies a space-variant blur, implemented via a multi-level Gaussian pyramid blended according to distance from the fixation point. This yields high fidelity in the central (foveal) region, with progressively increasing blur in the periphery.  The parameters of the foveation are taken from the implementation of ~\cite{jiang2015salicon} \footnote{Python Implementation: https://github.com/ouyangzhibo/Image\_Foveation\_Python/tree/master}. To capture variability in VLM output, we sample $N=5$ descriptions per image (both original and foveated) via multinomial sampling using the same VLM configuration.


\subsection{F-SUM Construction}\label{sec:F-SUM}

To measure understanding given the variety of possible fixations, we compute the semantic similarity between each foveated scene description and the gold standard (i.e., descriptions from the original, unfoveated image). We embed all descriptions using a text embedding model \citep{ zhang2024jasper,lee2025gemini} and for each gaze location, we calculate the mean pairwise cosine similarity between the $N$ gold standard embeddings and the $N$ embeddings from the corresponding foveated image.

Formally, the F-SUM is defined as a 2D matrix $M$, where each entry $M_{m,n}$ corresponds to a grid location $(m,n)$ and is computed as:
\begin{equation}
M_{m,n} = \frac{1}{N^2} \sum_{i=1}^{N}\sum_{j=1}^{N} \frac{{g}_i \cdot {f}_{m,n,j}}{|{g}_i| , |{f}_{m,n,j}|}
\end{equation}
Here, ${g}_i$ denotes the embedding of the $i$-th gold standard description, and ${f}_{m,n,j}$ denotes the embedding of the $j$-th foveated description at location $(m,n)$.

To enable comparison across scenes, we normalize $M$ to the $[0,1]$ range using the global distribution of mean pairwise similarities. Higher values in the F-SUM indicate gaze locations yielding descriptions more semantically aligned with the full-scene understanding; lower values indicate less informative fixations.





\subsection{Evaluation Metrics: Weighted Ripley’s K-function}\label{sec:metrics}

To quantify scene difficulty based on the image-computable F-SUM, we introduce a minimalist spatial metric inspired by Ripley’s K-function \citep{ripley1976second}, which is a spatial statistical measure used to analyze the distribution patterns of point data over space. A higher value of Ripley’s K-function at a given distance $r$ indicates a greater degree of spatial clustering at that scale. We adapted it to operate on weighted spatial data, i.e., weighted Ripley's K. The design of weighted Ripley's K is motivated by two criteria. All else being equal:

\begin{itemize}
    \item Global informativeness: Scenes with low (high) understanding values at all possible fixation points should receive a low (high) understanding score. 
    \item Spatial concentration: Scenes in which semantic understanding is spatially clustered should score higher than scenes that have such information spatially dispersed. This follows the intuition that scenes with spatially distributed information require multiple eye movements and longer exploration time to comprehend \citep{sweller2011cognitive}.
\end{itemize}




The weighted Ripley's K-function $K(r)$ for a given distance $r$ is calculated as:
\begin{equation}
K(r) = \frac{1}{N} \sum_{{p, q \in \mathcal{P}, p \neq q}} w_p w_q \cdot \mathbf{1}(r-1 \leq d_{pq} \leq r)
\end{equation}
Consider the F-SUM as a 2D matrix, where $\mathcal{P}$ be the set of coordinates of elements in the matrix, $p$ and $q$ are two pixels in $\mathcal{P}$, with coordinates $(i_p, j_p)$ and $(i_q, j_q)$. $d_{pq}$ is the Euclidean distance between $p$ and $q$. $w_p$ and $w_q$ are the values of pixels $p$ and $q$. $K(r)$ will be large if elements with higher value are clustering together within the distance $(r-1, r)$.


The final weighted Ripley's K-function is a weighted sum of $K(r)$. As $K(r)$ takes the value of each elements into account, this already satisfy the requirement for global informativeness. To further consider the spatial concentration, the weight should be inversely proportional to the distance. Let $K(r)$ be the weighted Ripley's K-function value at distance $r$. Let $w_r = \frac{1}{r}$ be the weight for distance $r$. Let $R$ be the maximum distance (we applied $R=10$ in our experiments). Then, the weighted Ripley's K-function score $S$ is calculated as:
\begin{equation}
S = \frac{\sum_{r=1}^{R} K(r) \cdot w_r}{\sum_{r=1}^{R} w_r}
\end{equation}
$S$ is further normalized to range from 0 to 1; a higher value means the scene is more difficult to understand.


\section{Experiments \& Results}

\subsection{Datasets}

Our image dataset comprises 277 images, including visual scenes from MSCOCO \citep{lin2014microsoft} as well as in-house photographed images depicting social interactions and complex human behaviors.  We selected this dataset to capture a range of scene complexities. Human annotated descriptions for each image were provided by five research assistants, who were instructed to describe each scene as accurately as possible without time constraints.

\subsection{Human Psychophysics}\

We conducted two psychophysical studies: (1) scene description response times and (2) scene descriptions under saccade-limited viewing using an EyeLink\textsuperscript{\textregistered} eye tracker. Each experiment included 277 trials, one per image. We tested whether the F-SUM score reflects scene comprehension difficulty. Using (1), we correlated F-SUM with response times and saccade counts while in (2), we correlated F-SUM with description accuracy---measured as similarity to human ground-truth annotations---under constrained viewing. Fig.~\ref{fig:Psychophysics} illustrates the trial timeline:
\begin{figure}[ht]
    \centering
    \includegraphics[width=0.8\linewidth]{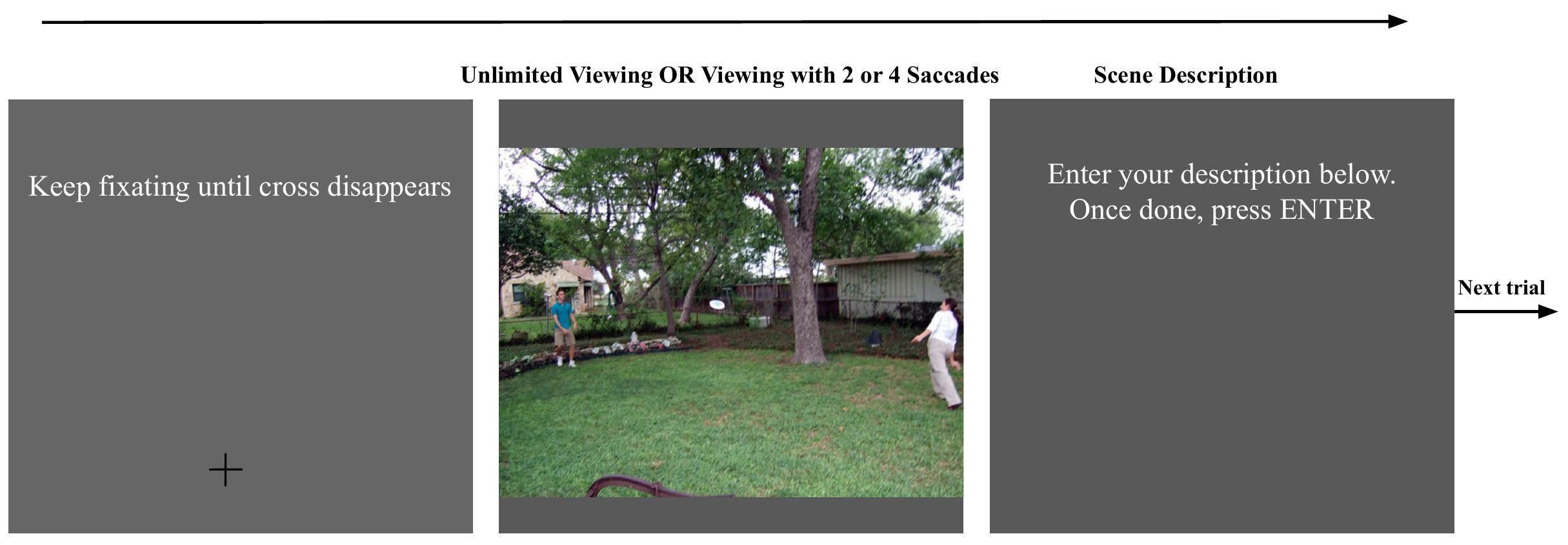}
    \caption{Overview of human psychophysics. For response time study (1), scenes were presented, participants explored the scene, and pressed the spacebar as soon as they determined that they comprehend the scene.  In the the constrained saccade allotment study (2), scenes were presented for a brief time and disappeared after the participants executed 2 or 4 saccades. Participants' initial fixation for both studies was a cross toward the bottom of the image.}  
    \label{fig:Psychophysics}
\end{figure}

\subsubsection{Response Time  Study}\label{sec:study1}

Participants ($N=17$) were presented with a scene and allowed to view it for an unlimited amount of time. They were instructed to press the spacebar on the keyboard as soon as they could describe the scene. Following this, participants were asked to type the description of the scene. To ensure response validity and prevent participants from circumventing the task, we required that the language embedding of each response be sufficiently similar to at least one of the gold standard annotated descriptions for the scene. Trials in which a participant's description had low similarity (Gemini embedding's cosine similarity less than $.75$) with all of the gold standards were discarded ($\sim$4\% of all trials). We recorded the response time (i.e., the interval between scene onset and spacebar press) and simultaneously collected eye-tracking data, including the total number of saccades.

\subsubsection{Saccade-Limited Image Presentation Study}\label{sec:study2}

In the limited saccades study, each scene was displayed for a restricted number of eye movements, either 2 or 4 saccades. The scene would be removed from the screen upon detection of the start of the 3rd or 5th saccade, respectively. Participants ($N=16$) were then instructed to provide a detailed description of the scene based on what they observed. Participants were randomly assigned to two groups (Group A and Group B), and the image set was divided into two subsets of equal size. Group A viewed the first subset of images under the 2-saccade condition and the second subset under the 4-saccade condition. Group B experienced the reverse assignment, such that each image subset was evaluated under both saccade constraints across participants.   The 2- and 4-saccade trials were randomly interleaved for each participant.

\subsection{Baseline Comparison Metrics}\label{metrics}

To our knowledge, no existing image-computable models directly predict the difficulty of real-world scene comprehension from the perspective of human observers, who experience spatial constraints such as foveated vision. However, several baseline metrics have been proposed to estimate related properties such as image clutter, visual complexity, and scene ambiguity. These metrics provide useful points of comparison for evaluating the F-SUM score. Notably, none of these baselines account for human visual foveation or its interaction with task-relevant information during scene understanding. Below, we detail the baseline metrics used for comparison.

\subsubsection{Clutter Metrics}
Visual clutter refers to the density of edges, texture, features, and objects, leading to performance degradation in perceptual tasks such as detection, identification, and visual search. Clutter is detrimental to perception in the visual periphery. Here we choose two traditional metrics, feature congestion  \citep{rosenholtz2005feature} and Subband Entropy  \citep{rosenholtz2007measuring}, as baselines to compare with our method.  If visual clutter is the main bottleneck of scene understanding, then these metrics should correlate with response times and the number of eye movements.

\subsubsection{Visual Complexity}
Image Complexity (IC)---also referred to as Visual Complexity---usually defined as the degree of disorganization and variety within an image, can serve as a possible metric for measuring the difficulty of understanding the scene. Low-level entropy measures or clutter metrics can capture some, but not all, aspects of visual complexity. For example, an image with random noise will have high entropy but can be meaningless to humans. Recent studies use deep learning models \citep{tudor2016hard, saraee2020visual, kyle2022predicting, kyle2023characterising, feng2022ic9600} to predict the complexity of images based on human-labeled datasets. There are also computational methods that calculate image complexity based on algorithmic information theory \citep{kolmogorov1965three, mahon2024minimum}. Based on the premise that a visual scene with more visual complexity could be more difficult to understand,  we included ICNet \citep{feng2022ic9600} and Meaningful Image Complexity \citep{mahon2024minimum} as comparators to F-SUM.


\subsubsection{Language Entropy}

Applying a VLM to generate scene descriptions can be framed as a specific instance of Visual Question Answering (VQA)---namely, responding to the implicit inquiry: ``Can you describe the scene?'' One way to quantify the difficulty or uncertainty of this task is by measuring the model’s predictive entropy over possible language responses~\citep{malinin2020uncertainty}:
\begin{equation}
    H(Y \mid x) = \mathbb{E}_{Y \sim p(y \mid x)}\left[-\log p(Y \mid x)\right] 
\end{equation}
\begin{equation}
    p(\mathbf{y} \mid \mathbf{x}) = \prod_{l=1}^{L} P(y_l \mid {y}_{<l}, \mathbf{x})
\end{equation}
where $x$ is the input image, $Y$ is the random output sequence, and $y_l$ is the $l$-th token in the sequence.

Because enumerating all possible output sequences is intractable, we approximate the predictive entropy via Monte Carlo sampling with $N$ generated outputs:
 \begin{equation}
     \hat{H}(Y) \approx - \frac{1}{N} \sum_{i=1}^{N} \log p(y_i \mid x)
\end{equation}
Since token-level log-likelihoods were not accessible from the closed-source model, we used the open-source VLM, Ovis2-8B~\citep{lu2024ovis} to compute entropy estimates\footnote{Ovis2-8B was selected based on its performance and model size on the \href{https://huggingface.co/spaces/opencompass/open_vlm_leaderboard}{OpenVLM Leaderboard}.}. For consistency, both language entropy and F-SUM scores were derived using this same model. Details of the sampling procedure are provided in the appendix.


\subsubsection{Directly Prompting a VLM}

Considering the fast evolution of current VLMs, we also directly asked a VLM to score the human difficulty in comprehending each image. Temperature is set to 0 to obtain deterministic results. Given the following prompt and a specific scene, we asked the VLM to return a score ranging from 0 to 1 to measure the human difficulty in describing the scene:

\begin{quote}
Please rate how difficult it would be for a human to describe this scene on a scale from 0.000 to 1.000, where 0.000 is very easy and 1.000 is very difficult. Respond with only a number between 0.000 and 1.000.    
\end{quote}

\subsection{Predicting Human Response Time and Number of Saccades}

We calculated the Pearson correlation coefficient ($r$) for F-SUM plus each of the metrics detailed in \ref{metrics} with two behavioral measures: human response time (RT) and the number of saccades made during scene comprehension. For our F-SUM score, we considered two VLM backends: 
\begin{itemize}
    \item \textit{F-SUM Score (closed-source)}, which utilizes GPT-4o to generate scene descriptions and employs the Gemini embedding model \footnote{For our experiments, we applied models/embedding-001} to extract embeddings for descriptions
    \item \textit{F-SUM Score (open-source)}, a fully open-source and smaller alternative that generates descriptions using Ovis2-8B \citep{lu2024ovis} and a distilled embedding model \footnote{Model name: stella\_en\_400M\_v5} \citep{zhang2024jasper} as the language embedding model. 
\end{itemize}

As shown in Tables~\ref{tab:metric_comparison_RT} and~\ref{tab:metric_comparison_Sacca}, we compared both versions of F-SUM against baseline metrics in their ability to predict response time (RT) and the number of saccades. Human–human correlation refers to the correlation between an individual participant’s behavioral data (i.e., response time or number of saccades) and the average behavioral data of the other 16 participants. The number of saccades measure exhibited higher internal correlation than RT.


To assess significance, we applied bootstrapping with $n=10{,}000$ across all images and participants. Results indicate that the F-SUM Score significantly outperforms all baseline metrics in predicting both response time and saccade count. Among the baselines, image complexity and language entropy also showed significant—but weaker—correlations with behavioral measures.

\begin{figure}[ht]
    \centering
    \includegraphics[width=0.9\linewidth]{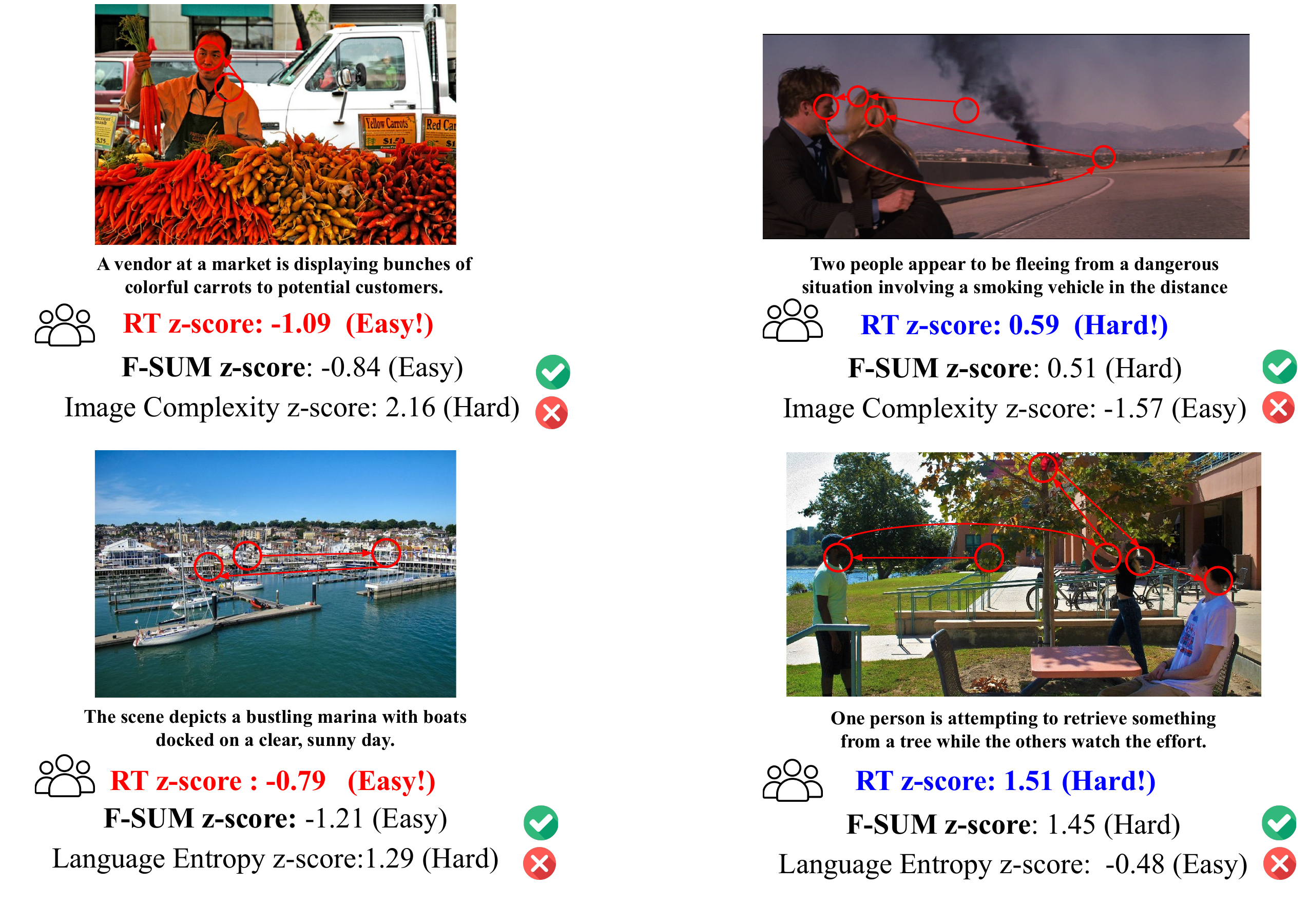}
    \caption{The difference between what F-SUM could capture and what Image Complexity and Language Entropy could capture. Both Image Complexity and Language Entropy fail to predict the response time of humans on some of the images. Gold standard of the scene and human eye movement scanpath are also shown here.}
    \label{fig:results_overview}
\end{figure}

To illustrate key differences between F-SUM and baseline metrics such as image complexity and language entropy, we present examples in Fig.~\ref{fig:results_overview}. These cases highlight instances where F-SUM provides a distinct assessment. For example, image complexity can be biased by the sheer number of visual elements---such as the densely packed carrots in the upper-left image---which may not reflect true cognitive demands.

Language entropy captures the diversity of sampled descriptions but may miss the underlying effort required to generate them. In the bottom-left example, captions consistently mention a person reaching into a tree while others look on. Although the descriptions seem coherent and varied, they require multiple eye movements to integrate dispersed visual information---an aspect that language entropy fails to consider but F-SUM captures.

\begin{table}[ht]
\centering
\caption{Correlation of each metric with the RT.  Human–human $r$ reflects leave-one-out correlation: for each participant, we computed $r$ between their RT and the mean of the other 16 participants, then averaged across all 17 participants. }
\label{tab:metric_comparison_RT}
\begin{tabular}{lccc}
\toprule
Metric & $r$ (95\% CI) &  $p$-value ($\Delta_m r>0$) \\
\midrule
\textbf{F-SUM Score (closed-source)}     & \textbf{0.47 (0.37 -- 0.56)} & ---              \\
\textbf{F-SUM Score (open-source)}     & \textbf{0.44 (0.34 -- 0.53)}& ---              \\
\midrule
Image Complexity (ICNet)       &   0.37 (0.26 -- 0.47)      &  0.03   \\
Language Entropy       &   0.36 (0.25 -- 0.46)     &  0.03    \\
GPT4o Prompted QA         &   0.22 (0.11 -- 0.33)             &   $< 0.001$  \\
Feature Congestion          &  -0.03 (-0.07 -- 0.17)            &  $< 0.001$  \\
Subband Entropy         & -0.01 (-0.12 -- 0.11)    &   $< 0.001$        &   \\
Meaningful Image Complexity         &   -0.03 (-0.14 -- 0.09)       &   $< 0.001$   \\
\midrule
Human-Human Correlation& 0.22 (---)  & --- \\    

\bottomrule

\end{tabular}

\footnotesize
\textbf{Notes:} For comparison metric $m$, $\Delta_m r = r_{F-SUM} - r_{m}$. The $p$-value ($\Delta_m r>0$) tests whether $\Delta_m r>0$ via bootstrap resampling. All metrics are compared to \emph{F-SUM Score (closed-source)} with the exception of Language Entropy, which is compared to \emph{F-SUM Score (open-source)} 
\end{table}








\begin{table}[ht]
\caption{Correlation of each metric with the number of saccades.  Human–human $r$ reflects leave-one-out correlation: for each participant, we computed $r$ between their saccade counts and the mean of the other 16 participants, then averaged across all 17 participants. \\
}
\begin{center}

\begin{tabular}{lcccccc}
\toprule
Metric & $r$ (95\% CI)  &  $p$-value ($\Delta_m r>0$) \\
\midrule
\textbf{F-SUM Score (closed-source)}     & \textbf{0.51 (0.42 -- 0.59)} &  ---             \\
\textbf{F-SUM Score (open-source)}     & \textbf{0.46 (0.36 -- 0.55)}  &   ---              \\  
\midrule
Image Complexity (ICNet)         &  0.39 (0.29 -- 0.49)           &  0.04   \\
Language Entropy         &  0.31 (0.20 -- 0.41)       &   $< 0.001$   \\
GPT4o Prompted QA       &   0.20 (0.08 -- 0.31)           &   $< 0.001$   \\
Feature Congestion          &   0.05 (-0.07 -- 0.17)       &   $< 0.001$   \\
Meaningful Image Complexity         &  0.04 (-0.08 -- 0.16)         &  $< 0.001$   \\
Subband Entropy         &  -0.01 (-0.13 -- 0.11)           &  $< 0.001$ \\
\midrule
Human-Human Correlation   & 0.43 (---) &   --- \\  
\bottomrule
\end{tabular}
\end{center}
\label{tab:metric_comparison_Sacca}
\footnotesize
\textbf{Notes:} For comparison metric $m$, $\Delta_m r = r_{F-SUM} - r_{m}$. The $p$-value ($\Delta_m r>0$) tests whether $\Delta_m r>0$ via bootstrap resampling. All metrics are compared to \emph{F-SUM Score (closed-source)} with the exception of Language Entropy, which is compared to \emph{F-SUM Score (open-source)}

\end{table}





\subsection{Predicting Scene Description Accuracy for Saccade-limited Image Presentations}

In addition to studying timing and eye movement, we assessed scene comprehension by evaluating the semantic accuracy of participants’ descriptions under time-restricted viewing conditions. Accuracy was defined as the average embedding similarity between each participant’s response (from Section~\ref{sec:study2}) and five human-annotated reference captions. Analyses were conducted separately for the 2-saccade and 4-saccade conditions.

As shown in Table~\ref{tab:metric_comparison_Sacca}, we computed the $r$ between each metric and the embedding-based accuracy scores separately for the 2-saccade and 4-saccade conditions. This allowed us to assess how well each metric, including F-SUM and baselines, predicted participants’ ability to extract and articulate the gist of a scene when access to visual information was limited by spatial viewing constraints.

\begin{table}[ht]
\centering
\caption{Correlation of each metric with the similarity of human scene description to gold standard description for saccade-limited presentation (2 or 4 saccades)}
\label{tab:metric_comparison_acc}
\resizebox{\linewidth}{!}{
\begin{tabular}{lcccc}
\toprule
Metric & $r_{\text 2-saccades}$ (95\% CI)  & $r_{\text 4-saccades}$ (95\% CI) \\
\midrule
\textbf{F-SUM Score(closed-source)}     & \textbf{-0.56 (-0.63 -- -0.48)}  & \textbf{-0.54(-0.62 -- -0.45)}       \\
\textbf{F-SUM Score (open-source)}     & \textbf{-0.53 (-0.61 -- -0.44)}  &  \textbf{-0.51(-0.59 -- -0.42)}         \\
\midrule
Language Entropy        &  -0.31(-0.41 -- -0.20)        &   -0.37(-0.47 -- -0.27) \\
Image Complexity (ICNet)         &  -0.20 (-0.31 -- -0.09)          &  -0.13(-0.24-- 0.01)  \\
GPT4o Prompted QA         &   -0.16 (-0.27-- -0.04)   &  -0.22(-0.33 -- -0.11) \\
Feature Congestion       &   0.07 (-0.05 -- 0.18)        &  0.01 (-0.11 -- 0.13) & \\

Subband Entropy       &    0.09 (-0.03 -- 0.20)         &    0.05(-0.07 -- 0.17)  \\

Meaningful Image Complexity         &  0.08 (-0.04 -- 0.19)         &  0.08(-0.04 -- 0.19)  \\
\bottomrule
\end{tabular}
}




\footnotesize
\textbf{Notes:} F-SUM score is significantly more correlated with the similarity of human response to gold standard compared with all other metrics ($p<.0001$, bootstrap). All metrics are comparing with \emph{F-SUM Score (closed-source)} except for Language Entropy, which is comparing with \emph{F-SUM Score (open-source)} 
\end{table}

\section{Discussion}

Previous work has proposed image-computable models to predict search times during visual search or object detection.  Here, we present a model metric to predict the time to comprehend a scene and show that it results in significantly higher correlation than other image clutter~\citep{rosenholtz2005feature,rosenholtz2007measuring}, complexity~\citep{feng2022ic9600,mahon2024minimum}, or language entropy metrics~\citep{malinin2020uncertainty}.   We also evaluated a direct prompting of a VLM model (GPT4o) to estimate RTs for the images, which resulted in low correlations. 

Our modeling approach, F-SUM score, takes into account the information critical to scene understanding (VLM component) and its interaction with the foveated human visual system (foveation pre-processing of images).  The Weighted Ripley’s K-function combines values across the F-SUM by penalizing distance across informative locations to reflect increasing time costs to fixate distant locations in the images.   Unlike the F-SUM score, the image clutter and image complexity metrics do not take into consideration which parts of the images contribute to the understanding of the scene, nor the interaction with the foveated visual system.   The language entropy measure is a proxy to difficulty in understanding a scene related to ambiguity in the scenes (i.e., objects or people in the scenes being occluded) but also does not take into account the spatial distribution of the scene understanding relevant information, nor its interaction with foveation.   Thus, aside from proposing a new metric, the work demonstrates that the main bottleneck for humans to understand scenes is whether the critical information is spread out or clustered in an image and whether it is identifiable and accessible in the visual periphery. 

There are limitations to our approach.  The F-SUM model does not explicitly predict response times, unlike approaches that compute information accrual dynamically to predict the time to reach a decision boundary \citep{rafiei2024neural}.   The pros of the F-SUM score approach are that there are no fitting parameters to the model, no training, and no requirements to use eye-tracking data.  This makes the model easy to compute. 

Humans differ in their ability to detect objects in the visual periphery, and thus, incorporating participant-specific peripheral properties might improve the model's ability to predict individual observers' response times. Finally, we did not explore the combination of various metrics (F-SUM score, language entropy, clutter, etc.), which might improve the ability to predict RTs.


\section{Acknowledgments}
This work is supported by the The Institute for Collaborative Biotechnologies through US Army Research Office Contract W911NF-19-D-001.

\bibliographystyle{elsarticle-harv}
\bibliography{ref}


\appendix
\section{Appendices}





\subsection{Usage of VLM and embedding model in F-SUM}

A VLM was applied to get the description for the unfoveated scene or the foveated scene. We sampled 5 descriptions for each foveated scene and unfoveated scene using the following prompt: 

\begin{quote}
    Make your best guess of what might be happening in this scene in one sentence. Avoid mentioning objects that do not aid in understanding the context of the scene.
\end{quote}

In F-SUM(closed-source), we used the API call of OpenAI to obtain the generation results of gpt4o model, using the default sampling parameter (temperature $= 1$). We obtained all the descriptions. We used the API call of Gemini  (models/embedding-001) to get the language embeddings. In F-SUM (open-source), we ran all the computations with a single NVIDIA GeForce RTX 4090 GPU. We used the default generation configuration of Ovis-8B: multinomial sampling with temperature $= 0.7$. Then we used a tiny open-source embedding model: stella\_en\_400M\_v5 to obtain the embedding for the descriptions.

\subsection{Configuration for calculating Language Entropy}

To make a meaningful comparison, controlling for the VLM model, the language entropy was calculated using Ovis-8B with the same sampling configuration. 10 Samples were generated to calculate the language entropy. We modified the implementation from \cite{kuhn2023semantic} to fit a vision language model. The prompt is the same as the prompt we used to get the scene descriptions. Models are loaded with Hugging Face
Library.


\subsection{Human Psychophysics experiments details}

The materials and procedures used in this study were approved by UCSB's Internal Review Board. For the response time study, we have 17 undergraduate students (1 male, 16 female, aged: 19-21 (19.9 ± 0.7) years). For the saccade-limited presentation study,  we recruited 16 undergraduate students (4 male, 12 female, aged 19-22 (20.5 ± 1.1) years). Subjects were recruited to participate in the experiment for research credit. All participants signed a consent form to provide informed consent to participate in the study. Participants had normal or corrected-to-normal vision.




\end{document}